\documentclass[letterpaper, 10 pt, conference]{ieeeconf}  
\IEEEoverridecommandlockouts                              

\usepackage{mathptmx} 
\usepackage{amsmath} 
\usepackage{amssymb}  
\usepackage{mathtools}
\usepackage[linesnumbered]{algorithm2e}
\DeclareMathOperator*{\argmax}{argmax}
\usepackage{graphicx}
\usepackage{caption}
\usepackage{subcaption}
\usepackage{color,soul}
\usepackage{balance}
\title{\LARGE \bf
 MeSA-DRL: Memory-Enhanced Deep Reinforcement Learning for Advanced Socially Aware Robot Navigation in Crowded Environments
}

\author{ Mannan Saeed Muhammad$^{1,*}$, Estrella Montero$^{2}$
\thanks{*This work was supported by National Research Foundation of Korea (NRF) funded by the Government of Korea (MSIT) (No. NRF-2022R1G1A1012746)}
\thanks{$^{1}$Mannan Saeed Muhammad is with the Department of AI \& Robotics, Sejong University, South Korea,
        {\tt\small mannan@sejong.ac.kr}}%
\thanks{$^{2}$Estrella Montero is with the Department of Electrical and Computer Engineering, Sungkyunkwan University, South Korea,}%
}

\begin{document}
\maketitle
\thispagestyle{empty}
\pagestyle{empty}

\begin{abstract}
Autonomous navigation capabilities play a critical role in service robots operating in environments where human interactions are pivotal, due to the dynamic and unpredictable nature of these environments. However, the variability in human behavior presents a substantial challenge for robots in predicting and anticipating movements, particularly in crowded scenarios.
To address this issue, a memory-enabled deep reinforcement learning framework is proposed for autonomous robot navigation in diverse pedestrian scenarios. 
The proposed framework leverages long-term memory to retain essential information about the surroundings and model sequential dependencies effectively.  
The importance of human-robot interactions is also encoded to assign higher attention to these interactions. 
A global planning mechanism is incorporated into the memory-enabled architecture.  
Additionally, a multi-term reward system is designed to prioritize and encourage long-sighted robot behaviors by incorporating dynamic warning zones. 
Simultaneously, it promotes smooth trajectories and minimizes the time taken to reach the robot's desired goal. 
Extensive simulation experiments show that the suggested approach outperforms representative state-of-the-art methods, showcasing its ability to a navigation efficiency and safety in real-world scenarios.

\end{abstract}

\section{INTRODUCTION}

Autonomous mobile robots have become a necessity due to factors such as an aging population, scarcity of workforce, and the growing preference for non-contact services \cite{kim2021preference}
Consequently, autonomous robots must not only ensure their safe movement in static environments but also demonstrate the ability to navigate through densely crowded human spaces \cite{kim2021preference}.
This poses a significant challenge for the robot due to the intricate nature of predicting and modeling human intentions and preferred movement patterns in crowded environments  \cite{xiao2022motion}.
The integration of social cues and awareness into robot navigation has been proposed as a means of alleviating some of the difficulties associated with human-robot interactions.
These navigation algorithms with a social awareness factor consider elements like proximity, social norms, and intention estimation \cite{fan2019getting,bonny2022highly}.
However, these approaches often rely on specific rules and heuristics, thus limiting their adaptability when applied to diverse scenarios.

The progress in deep reinforcement learning (DRL) has been utilized as an intelligent decision-making approach for managing navigation tasks in robotics \cite{han2022reinforcement, everett2021collision, zhu2022hierarchical}.
In this framework, the sensor observations assume the role of states for the agents involved \cite{cimurs2021goal}, including positions and velocities, through deep neural networks (DNNs) \cite{chen2017decentralized, everett2018motion, chen2019crowd, liu2021decentralized}. The DRL methodology identifies the optimal policy for guiding the robot through interactions with its environment, achieving this by maximizing the cumulative reward obtained from its actions.    
However, certain DRL methods do not possess the flexibility to handle the variations in the number of humans in the crowd. Consequently, it becomes necessary to retain the DNN whenever there is an alteration in the number of people in the crowd \cite{chen2017decentralized}.
In addition, many DRL algorithms have limitations given the unpredictable nature of human behavior, where factors such as physical condition, ongoing activities, and individual size influence movement speed, and simulating these states accurately becomes a challenging task.
By utilizing robot egocentric perception, certain contributions have successfully achieved a level of human awareness \cite{singh2023behavior}. However, this progress introduces new challenges, including determining the appropriate duration for maintaining awareness of detected human presence, leading to predominantly reactive adaptations. Addressing this, navigation-based memory neural networks have emerged to tackle the issue of retaining information over longer periods. These networks are designed to monitor static objects \cite{reich2020memory}.

This study introduces an innovative memory-enhanced socially aware DRL (MeSA-DRL) algorithm that leverages human cognition's reliance on memory, enabling the robot to sustain acquired awareness and navigate safely through densely crowded environments. As a result of anticipating diverse human actions, this algorithm enhances robot trajectories, collision avoidance, and goal achievement. The main contributions include the memory-enabled DRL algorithm that incorporates gated recurrent units (GRU), emphasizes human-robot interaction, and integrates global planning. A multi-term reward system with warning zones further ensures safe navigation and trajectory efficiency. Experimental results validate the algorithm's significant improvements in safety and efficiency within dense crowds, surpassing previous methods.

\begin{figure*}[!h] 
      \centering
      \includegraphics[scale=0.37]{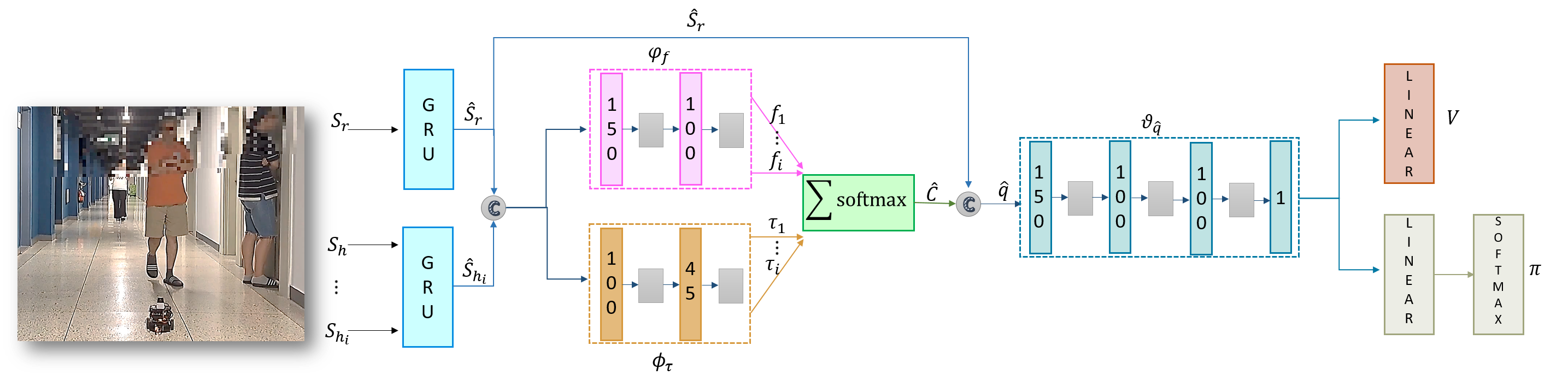}
       \caption{Schematic of the proposed network model. Pedestrian and robot state vectors are concatenated to form a pairwise combined state vector, with the network outputting policy $\pi$ over possible actions and the value $V$ of the current state.}
      \label{fig:neuralnetwork}
   \end{figure*}
\section{RELATED WORK}
In early phases, crowd navigation primarily relied on reaction-based motion planners \cite{van2008reciprocal, van2011reciprocal, ferrer2013robot}.
To achieve optimal collision-free navigation, methods such as Reciprocal Velocity Obstacle (RVO) \cite{van2008reciprocal} and Optimal Reciprocal Collision Avoidance (ORCA) \cite{van2011reciprocal} introduced geometric concepts that involved adjusting the velocity vector based on each agent's current position. 
Social Force (SF) \cite{ferrer2013robot} utilized attractive and repulsive forces to simulate the interactions between the robot and other agents. 
However, this type of reaction-based approach merely makes decisions based on the present state, failing to consider the potential long-term impacts of the current action and how future states might influence the current decision. 
In recent years, learning-based robot navigation techniques have been the focus of research, where an optimal strategy is learned through interaction with the environment, guided by long-term cumulative rewards \cite{fan2020distributed, perez2021robot, sathyamoorthy2020frozone}.

Some approaches involve decentralized collision avoidance techniques that use sensor data to guide the behavior of individual agents \cite{fan2020distributed, jin2020mapless}.
However, owing to the raw state inputs, these approaches require a substantial amount of data to establish the mapping relationship.

Collision avoidance with deep reinforcement learning (CADRL) \cite{chen2017decentralized} represents social navigation implementation using DRL.
Although CADRL's value function is successful in scenarios involving robots and pedestrians, it ignores social interaction among pedestrians.
Utilizing LSTM to encode both the robot's state and the states of all pedestrians, LSTM-RL \cite{everett2018motion} further augments the capabilities of CADRL.
The state-of-the-art advancements include socially aware RL (SARL) \cite{chen2019crowd}, which uses self-attention mechanisms, to capture and infer relationships between agents in the environment \cite{chen2020robot}.
Despite these advancements, ensuring human safety has remained a challenge attributed to reliance on complete knowledge of pedestrian information.
Considering heterogeneous characteristics, the DRL-based navigation method is extended to collision avoidance scenarios with heterogeneous decision-making agents \cite{zhu2022collision, samsani2021socially}.
Introducing a safety measure, dynamic warning zones around humans are established based on their size, speed, and step length ensuring collision-free navigation \cite{montero2023dynamic}.
A dynamic path planning approach for mobile robots is proposed, integrating a gated recurrent unit (GRU) \cite{yuan2019novel, zeng2019navigation}.
As mentioned previously, these studies have successfully tackled multi-agent navigation, yet they were unable to cater to the diversity of real-time scenarios that a robot can encounter, especially with increased crowd density. 
In unfamiliar environments, the robot struggles to plan a coherent and smooth path due to inherent precision requirements in DRL techniques for constructing accurate models.

This work introduces a comprehensive approach to enhance robot navigation in diverse pedestrian environments by employing a multi-term reward system that anticipates dynamic situations. It leverages a memory-enhanced DRL framework, an attention mechanism to assess individual significance and a global planning system to ensure safe and efficient navigation.


\section{METHODOLOGY}
In this section, a Markov Decision Process (MDP) is used to describe the autonomous navigation of robots. The MDP problem is addressed by defining the observations, employing the value-based method, and formulating a  multi-term reward. Subsequently, the derivation of the network architecture is described. 

\subsection{Problem Formulation}
This study considers a robot navigating among $n$ number of humans within a 2D plane. The robot and humans are modeled as a circle of radius $r_r$ and $r_{ih}$ respectively. Utilizing the observations including position and velocity, the robot establishes a state space $\mathbf{S}$ comprising both the robot's internal state $\mathbf{S}_r$ and the observable information of each human $\mathbf{S}_{ih}$ at time $t$.

To enhance the precision of the navigational analysis, a robot-centric coordinate frame is established, wherein all agents' positional and velocity vectors are transformed into coordinates centered around the robot. In this framework, the x-axis is oriented, extending from the robot's present location toward its intended destination. 
Subsequently, the states of both robot and human are defined as follows:
\begin{align} \label{Eq:roce}
   \mathbf{S}_r     &=[d_g,\mathbf{v}_r,r_r,v_{max}]_{t}, \nonumber\\ 
   \mathbf{S}_{ih}  &=[\mathbf{p}_{ih},\mathbf{v}_{ih},r_{ih},d_i,r_{ih}+r_r]_{t}, 
\end{align}
where $d_g$ denotes the distance from the robot to its goal $\mathbf{g}=[g_x, g_y]$, and $d_i$ is the distance from the robot's position to its $i$-th neighbor's position, $\mathbf{v}_r$ is the velocity of the robot, and $v_{max}$ indicates the preferred velocity for the robot. Subsequently, the position and velocity of $i$-th human is given by $\mathbf{p}_{ih}$, and $\mathbf{v}_{ih}$. 

At any time $t$, the state space is represented as the joint state $\mathbf{S}_j$, which integrates state information from both the robot and the human and is presented as follows:
\begin{equation} \label{Eq:js}
    \mathbf{S}_j = [\mathbf{S}_r, \mathbf{S}_{1h}, \mathbf{S}_{2h},..., \mathbf{S}_{ih}]_{t}
\end{equation}

The neural networks are structured to establish an end-to-end mapping function from the state space $\mathbf{S}$ to the value function $V$.  
Within the framework of value-based RL, environmental observations are employed to determine the best navigation strategy. The optimal navigation policy $\pi^{\ast}$ is then derived to obtain the optimal value function $V^\ast$ for the joint state $\mathbf{S}_j$ at time $t$.
\begin{equation}
    V^\ast(\mathbf{S}_j) =\sum\limits_{k=0}^{N-1} \left(R(^k\mathbf{S}_j,\pi^{\ast}(^k{S}_j)) \cdot \gamma^{k\Delta t|v_{max}|}\right) ,
    \label{Eq:value}
\end{equation}

where $N$ denotes the number of decisions from the initial state to the end state, $R(\cdot)$ represents the reward obtained at time $t$, $\Delta t$ is the time step, and $\gamma$ is the discount factor within the range of $0<\gamma\leq 1$. To ensure numerical stability, the preferred speed $v_{max}$ is utilized as a normalization parameter.

The robot moves to the next state based on an unknown state transition probability function P$(\mathbf{S}_{t+1}|\mathbf{S}_t, \alpha_t)$ at each time-step $\Delta t$ by performing an action $\alpha_t$ sampled from the learned policy $\pi$. 
The state value is estimated at each time step of an episode through Monte Carlo Value Estimation after the episode concludes. 
Following this, the best action plan is translated into the optimal policy $\pi^{\ast}$, as shown below:
\begin{equation}\label{Eq:policy}
    \pi^{\ast}(\mathbf{S}_j) = \underset{\alpha_t}{\argmax} \left(R(\mathbf{S}_j,\alpha_t) + V^\pi(\mathbf{S}_j)\gamma^{\Delta t |v_{max}|} \right).
\end{equation}
The value network is trained through the temporal difference method in conjunction with experience replay \cite{chen2019crowd}.

In reinforcement learning, the agent's objective is to maximize the expected reward $R(\mathbf{S}_j,\alpha_t)$.
To account for the challenges of forecasting unpredictable human behavior, a multi-term reward system is designed based on dynamic warning zones (wz) \cite{samsani2021socially,montero2023dynamic}.  
The robot is encouraged to maintain a safe distance from humans by steering clear of crossing the warning zones.
These dynamic warning zones simulate realistic behavior by defining circular sectors around humans influenced by their dimensions, speed, and length of their steps.
To shape the robot's behavior based on warning zones, the following reward is formulated:
\begin{equation}
   ^tR_{wz}=\left\{
  \begin{array}{@{}ll@{}}
    0.2 \left( e^{k_{wz}} - 0.3\right), & \text{if}\ \text{exceed wz} \\
    ^tR_{route}, & \text{otherwise}.
  \end{array}\right.
\label{Eq:rewardwz}  
\end{equation}

The penalty $^tR_{wz}$ is exponential and is controlled by the factor $k_{wz}=(d_{i}-r_{wz} -r_{ih})$, wherein $d_{i}$ represents the magnitude of $||\mathbf{p}_{r}-\mathbf{p}_{ih}||_2$ which is the distance between the robot and its neighbor, $r_{wz}$ denotes the radius of the dynamic warning zone, and $r_{ih}$ is the human radius as per respective human group.
To discourage the robot from entering these zones, it is punished if $d_i>r_{wz}-r_{ih}$.

Furthermore, to encourage the robot to move towards the goal while minimizing the time taken, an additional route reward $R_{route}$ is introduced, as follows:
\begin{equation}
   ^tR_{route}= 0.01\left(-{d}_{g}[t] + {d}_{g}[t-1] \right).
    \label{Eq:routerew}  
\end{equation}
Here, ${d}_{g}[t]$ and ${d}_{g}[t-1]$ are the magnitudes of $||\mathbf{p}_{r} - \mathbf{g}||_2$ corresponding to the distances between the robot's position and the goal at two consecutive time steps. 

Additionally, $R_{route}$ is utilized to encourage the robot to adopt trajectories with fewer irregularities and deviations.

Moreover, to aid the robot in maintaining a suitable distance from humans during navigation, a dedicated penalty for discomfort $^tR_{disc}$ is incorporated and is invoked if the robot approaches the human too closely i.e., 
\begin{equation}
^tR_{disc}=0.25(d_{i} - 0.3).
\end{equation}
The robot is awarded or punished for its performance during the navigation given by $^tR_{nav}$.  If it approaches its goal, i.e., $(p_x - g_x, p_y-g_y)=0$, the robot receives a high reward of 10. 
\begin{equation}
    ^tR_{nav}=\left\{
  \begin{array}{@{}ll@{}}
    10, & \text{if reached goal}\   \\
    -0.25, & \text{if collison.}\
  \end{array}\right.
\label{Eq:rewardnav}  
\end{equation}
The reward for timing out $^tR_{time}=-10$, is enforced when time $t$ surpasses the maximum length of episode $T$.

The cumulative multi-term reward at time $t$ is defined as follows:
\begin{equation}
    R(S_j,a_t) =  ^tR_{wz} + ^tR_{nav} + ^tR_{disc} + ^tR_{time}
\end{equation}

In this study, a memory-enabled mechanism is specifically constructed to predict $ V^\ast(\mathbf{S}_j)$ as indicated in (\ref{Eq:value}), followed by (\ref{Eq:policy}), allowing the robot to make more reliable navigation decisions.

\subsection{Network Architecture}
The scope of this study is focused on memory's role, investigating how working memory influences navigation performance. In order to achieve this, a GRU memory-enabled DRL framework referred to as MeSA-DRL is proposed, which helps the robot accumulate knowledge and preserve it as needed. MeSA-DRL is a memory-enhanced socially aware DRL for crowded environments.
The GRU model \cite{cho2014properties, chung2014empirical}, known for its proficiency in learning from sequences, is utilized due to its streamlined complexity compared to the traditional LSTM model, facilitating information flow regulation through gates. The overall network architecture is illustrated in Fig. \ref{fig:neuralnetwork}.

For preliminary feature extraction, at each time step, the input sequence of human states {$\mathbf{S}_{ih}$} is fed into the corresponding GRU. The GRU layer is configured with specific parameters such as the input size, hidden size, and bi-directionality. The input size is set to 7, representing the dimensionality of the human state, i.e., the number of features for each state, as defined in {(\ref{Eq:roce})}. The size of the hidden state, set to 20, represents the number of hidden units in the GRU layer and determines the dimensionalilty of the output. The hidden state acts as a memory that encodes information about the past context of the input sequences. Finally, the proposed method adapts the merits of a bi-directional GRU to handle the variable size of the input vector and store the relative information of the humans, where the information is processed simultaneously in both forward and backward directions.
The robot state $\mathbf{S}_r$ is individually passed through a separate GRU layer. 
This model utilizes the previous context of input sequences to later allow better understanding of the dynamic relationships within the environment.
The resultant robot vector $\mathbf{\hat{S}}_r$ is expanded to match the size of $\mathbf{\hat{S}}_h$. These outputs are then concatenated and input into a multilayered perception (MLP) represented as $\varphi(\cdot)$, to generate a human-robot interaction vector $\mathbf{f}_i$, as given below:
\begin{equation}\label{Eq:invect}
    \mathbf{f}_i = \varphi_f (\mathbf{\hat{S}}_r, \mathbf{\hat{S}}_{ih}), ~~i = 1,...,n~~
\end{equation}
Concurrently,  they are used in another MLP to compute attention weights, $\mathbf{\tau}_i$, of each human signifying their importance to the robot as shown in the following:
\begin{equation}\label{Eq:attewei}
    \mathbf{\tau}_i = \phi_{\tau} (\mathbf{\hat{S}}_r, \mathbf{\hat{S}}_{ih}), ~~i = 1,...,n~~
\end{equation}
A linear combination of the \textit{softmax} function applied to the attention weights by human-robot interaction vector is computed as the crowd feature $\mathbf{\hat{c}}$:

\begin{equation}
   \mathbf{\hat{c}} = \sum\limits_{i=1}^n \left(\text{softmax}(\mathbf{\tau}_i) \cdot \mathbf{f}_i\right)
\end{equation}
The crowd feature is concatenated with the robot vector $\mathbf{\hat{S}_r}$ and the result is fed into MLP $\vartheta(\cdot)$ with RELU activation to separate the estimation of the value function and the policy. 

Finally, $\mathbf{\hat{q}}$ is calculated as the input to a single fully connected layer to obtain the value function  $V(\mathbf{S}_j)$, and a softmax is  added to produce a categorical policy output $\pi(\mathbf{S}_j)$ as given by:
\begin{equation}
    \mathbf{\hat{q}} = \vartheta_{ \hat{q}} (\mathbf{\hat{c}}, \mathbf{\hat{S}}_r; W_{\hat{q}}).
\end{equation}
where $W_{\hat{q}}$ represents the network weights.

\RestyleAlgo{ruled} 
\begin{algorithm}[b] 
  \SetAlgoLined
  
  Initiate action space $\alpha$\;
  Initiate pre-trained value network $V$ \;
  Occupancy map is built $map$\;
  Robot's goal is set $\mathbf{g}$\;
  
  \BlankLine
  \While{$\mathbf{g}$ \textnormal{is not reached}}{
    Update the joint state of the system $\mathbf{S}_j$ \;
    Calculate a global path $G$ \;
    Select a sub-goal within a short distance $g_{sub}$ \;
    $\alpha_t \Leftarrow \argmax_{\alpha_t \in A} R(^k\mathbf{S}_j, \alpha_t)+ \gamma^{k\Delta t|\mathbf{v}_{max}|} V(\mathbf{S}_j)$\;
    Update $\mathbf{p}_r$ $\Leftarrow $ $g_{sub}$ \;
  }

  \caption{MeSA-DRL algorithm}\label{algorithm}
\end{algorithm}

\subsection{Global Planning Algorithm Design}
To achieve the objective of robot navigation, this study proposes integrating memory-enabled training within a navigation framework. This framework consists of a 2D occupancy grid, denoted as $map$, and employs a global planner such as Dijkstra's algorithm to generate an optimal global path from the robot's initial position to its desired goal $g$. Given the global path and the current position of the robot $\mathbf{p}_r$, a dynamic sub-goal $g_{sub}$ is selected, serving as a proximate target that facilitates the comprehensive task of navigation. This methodology enhances flexibility, particularly in scenarios populated with multiple unknown obstacles, such as when the robot diverges from the global path by a specified distance. The algorithm is described in Algorithm \ref{algorithm}.

\section{SIMULATION EXPERIMENTS}
This section outlines the design of the navigation scenarios, along with the training and testing protocols, complemented by a description of the evaluation metrics. Subsequently, the performance of the proposed method is validated through comparison with other state-of-the-art methods.

\subsection{Simulation Setup}

For the simulations, a 2D scenario is created where the robot maneuvers through various environmental settings. Contrasted with preceding work \cite{chen2019crowd}, the simulated environment encapsulates more authentic crowd navigation contexts. To mirror the complexities inherent in real-world crowd navigation, several scenarios have been formulated: one devoid of obstacles i.e., empty scenario,  and others containing ten and fifteen dynamic obstacles moving in groups. These scenarios are leveraged to identify whether the robot policies exhibit a freezing robot phenomenon and to evaluate the quality of its trajectory. Humans within the simulated environment are depicted as circles with radii $r_{ih} \in [0.2, 0.6]$m and random velocities $v_{ih}$ spanning $[0.5, 1.8]$m/sec, reflecting their continuously variable intentions and distinct traits \cite{montero2023dynamic}.
Additionally, reciprocal motion among humans is generated by ORCA \cite{van2011reciprocal}.
The consideration of diverse human motion is crucial for developing a more interactive human crowd and for assessing the navigational safety of the robot.

The proposed network is developed within the PyTorch deep learning framework \cite{paszke2019pytorch}. 
Initially, the model completes $3,000$ demonstration episodes utilizing ORCA {\cite{van2011reciprocal}}. Subsequently, it is trained for $400$ epochs through the gradient descent method (learning rate: $0.01$). Then, the robot undergoes training across $10,000$ episodes with a learning rate of $0.001$.
Considering the significance of future rewards, a discount factor $\gamma$ of $0.9$ is assumed.
The exploration rate of the greedy strategy reduces linearly from 0.5 to 0.1 in the first $4,000$ episodes, after which it stays constant for the remaining episodes. Subsequent to the initial $1,000$ episodes, mini-batches, each consisting of $100$ samples are utilized to augment the value network.
The robot is specified with a radius of $0.3$m and a peak velocity of $1$m/sec.
This investigation replicates four state-of-the-art navigation methodologies, specifically CADRL \cite{chen2017decentralized}, LSTM-RL \cite{everett2018motion}, SARL \cite{chen2019crowd}, and CAM-RL \cite{samsani2023memory}.

\subsection{Comparative Evaluation}
Navigation performance is evaluated across 500 test cases using the neural network weights preserved during the training phase.
\begin{figure}[!t] 
      \centering
      \includegraphics[scale=0.245]{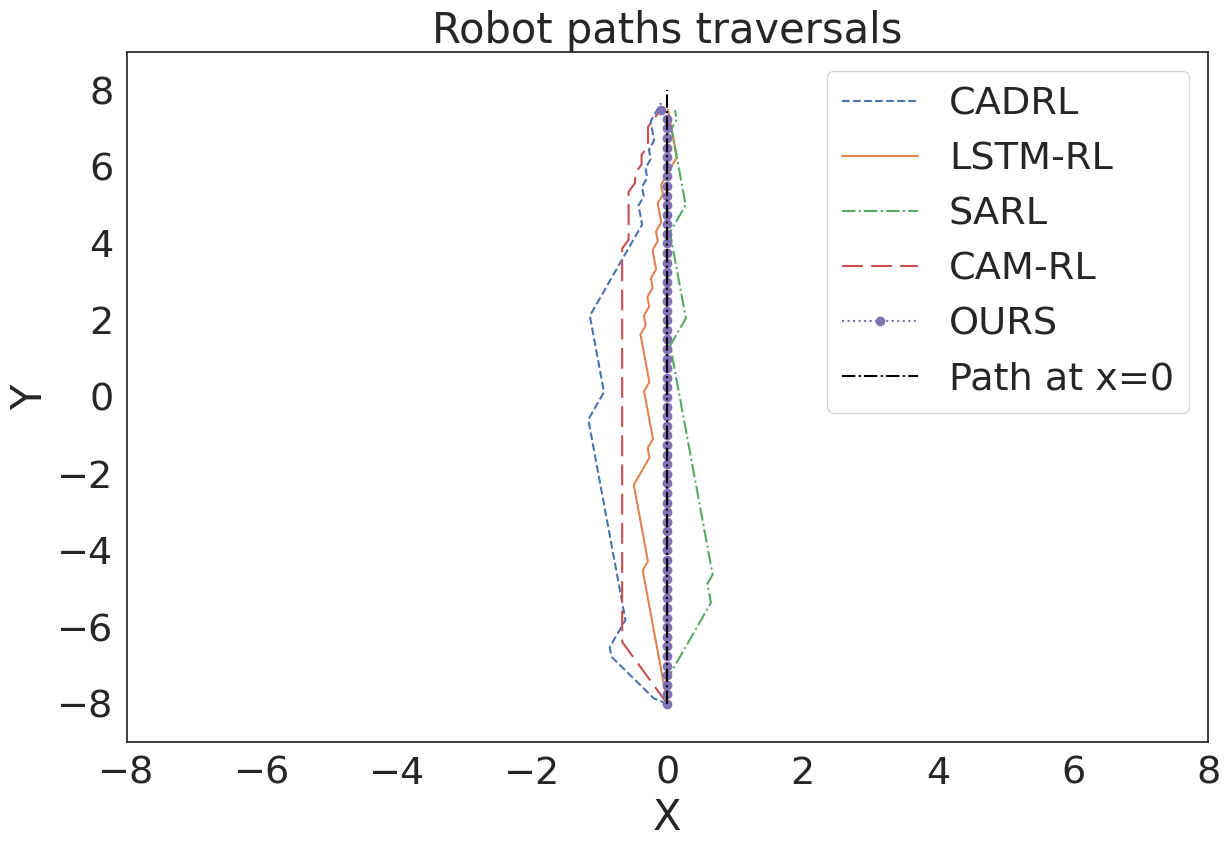}
      \caption{Illustration of the robot path traversals for each method in an obstacle-free space. }
      \label{fig:paths}
   \end{figure}

\begin{figure}[!t] 
      \centering
      \includegraphics[scale=0.24]{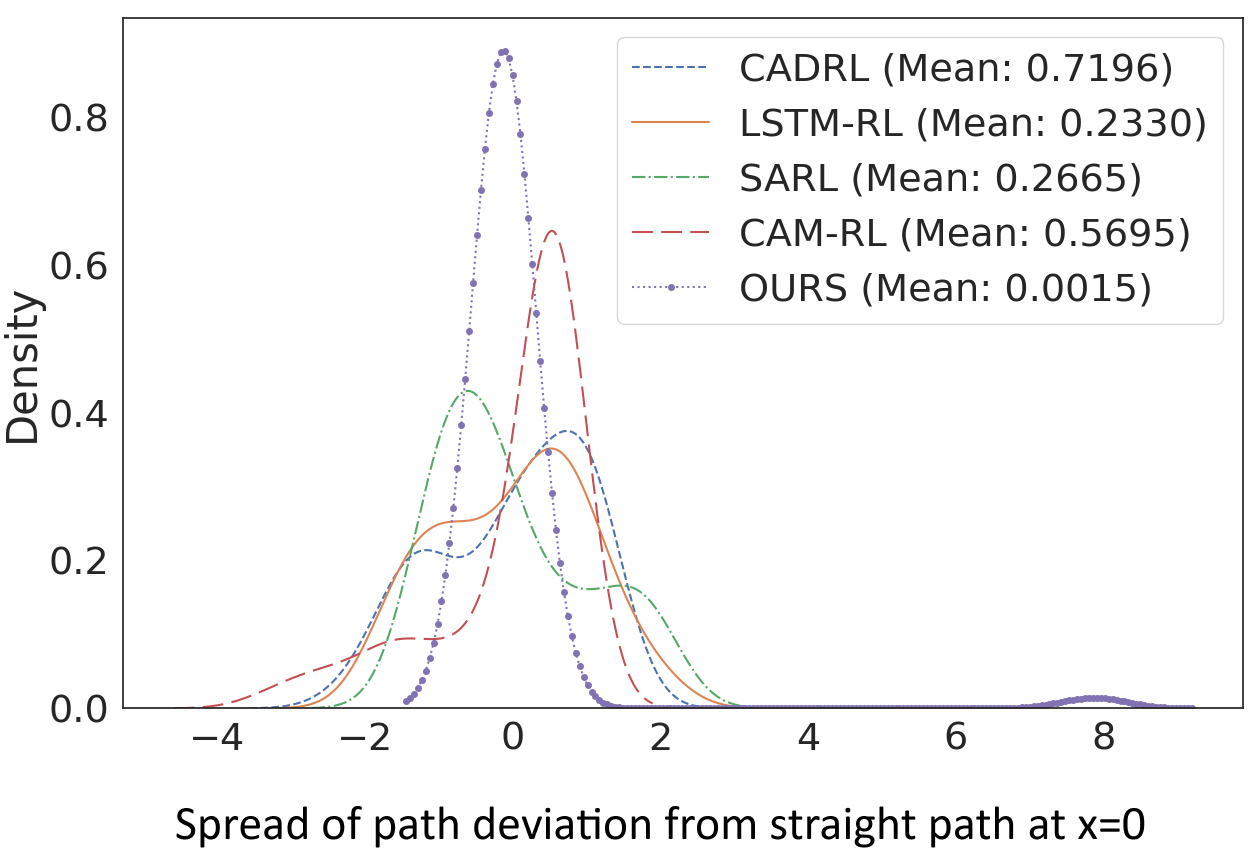}
     
       \caption{Spread of path deviation from the straight path at $x=0$ for each method's path traversal.}
      \label{fig:deviation}
   \end{figure}



\begin{figure*}[!t]
\centering
\begin{subfigure}{.3\textwidth}
  \centering
  \includegraphics[width=\textwidth]{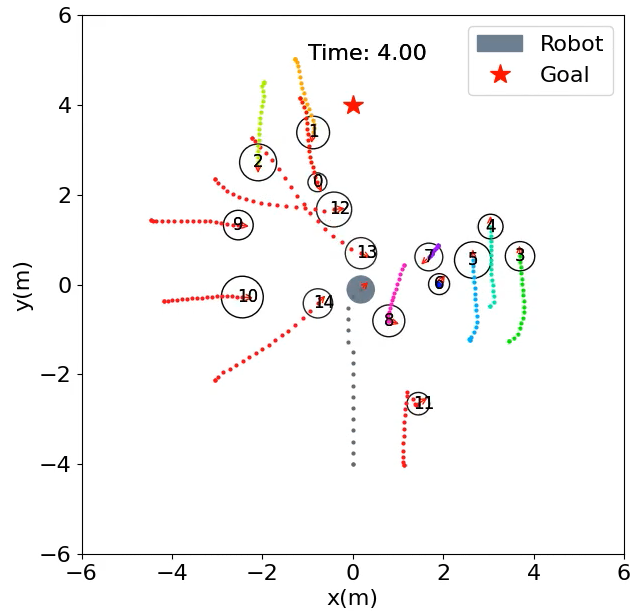}
  \caption{CADRL}
  \label{fig:cadrl_15h}
\end{subfigure}%
\hspace{0.2cm}
\begin{subfigure}{.3\textwidth}
  \centering
  \includegraphics[width=\textwidth]{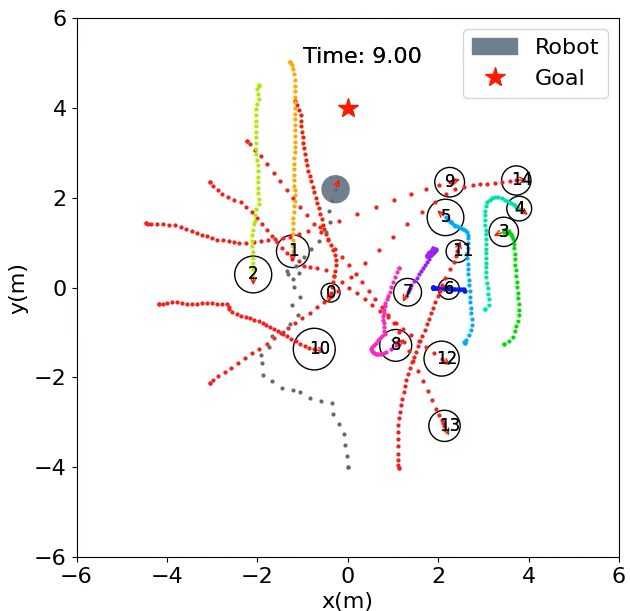}
  \caption{LSTM-RL}
  \label{fig:lstm_15h}
\end{subfigure}%
\hspace{0.2cm}
\begin{subfigure}{.298\textwidth}
  \centering
  \includegraphics[width=\textwidth]{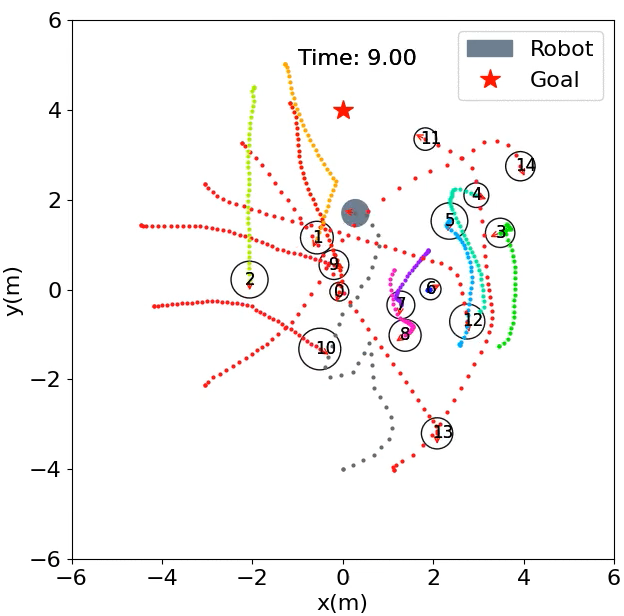}
  \caption{SARL}
  \label{fig:sarl_15h}
\end{subfigure}%
\hfill
\begin{subfigure}{.3\textwidth}
  \centering
  \includegraphics[width=\textwidth]{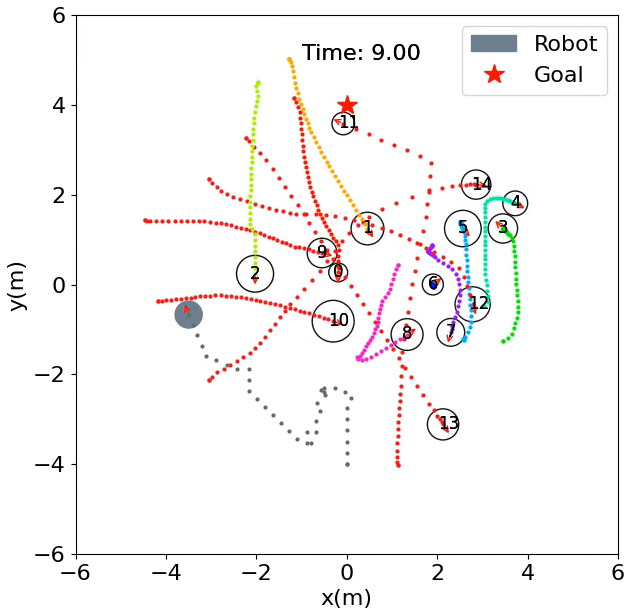}
  \caption{CAM-RL}
  \label{fig:camrl_15h}
\end{subfigure}%
\hspace{1cm}
\begin{subfigure}{.3\textwidth}
  \centering
  \includegraphics[width=\textwidth]{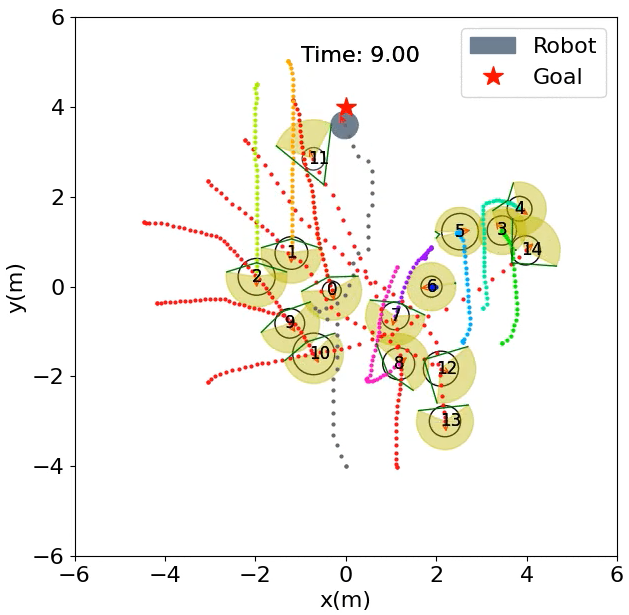}
  \caption{MeSA-DRL}
  \label{fig:our_15h}
\end{subfigure}%
\caption{Comparative analysis of local trajectories in a random test episode featuring ten humans within a grouped-human scenario. All experimental conditions maintain identical starting points, goal positions, and time steps.}
\label{fig:compa15h}
\end{figure*}

\textbf{Quantitative Results:} 
This section details a comprehensive evaluation comparison, spotlighting vital metrics in the context of robot navigation in dynamic environments. The success rate (SR) is defined as instances wherein the goal is attained without any collisions, while average time (NT) is the average duration for all successful navigation.
The average length of the path traveled for successful navigation (PL) is also considered to evaluate the efficiency of the navigation of the robot.
Additionally, average cumulative reward (AR) signifies that the robot must not only interpret human behavior but also actively interact with individuals. Failures are characterized by the collision rate (CR) between the robot and humans, as well as instances of time out (TO), which occur when the robot does not achieve its goal within the stipulated time frame.
The results of robot navigation performance revealed that the number of collision cases reduced to less than $5$ when the success rate exceeded $90\%$ as displayed in Table \ref{table:compa} and \ref{table:compa15}.

The efficacy of CADRL is considered suboptimal since it only accounts for pair interactions within a crowd, neglecting remaining interactions.
The LSTM-RL enhances the SR of navigation to $0.96$ with ten humans as it considers each pair of interaction information.
However, this method is not effective in extracting global information, which increases the navigation time to $22.15$ sec with fifteen humans, indicating that the strategy is not the most optimal.
In contrast, SARL, by analyzing the spatial relationship in the crowd, lowers navigation time and exhibits advanced results compared to CADRL, LSTM-RL, and CAM-RL, elevating SR to $0.97$ and maintaining a low CR of $0.03$. However, SARL tends to adopt a conservative approach, resulting in numerous occurrences of timeouts. The suggested method suppresses SARL in both navigation time and cumulative reward, demonstrating the efficacy of the proposed method.
Moreover, this method also outperforms CAM-RL in more challenging scenes, due to memory-based architecture, which incorporates a softmax layer to generate a categorical optimal strategy, while CAM-RL registers a prolonged NT (average time taken by the robot to reach the goal). 
A higher evaluation criterion, involving fifteen humans navigating both individually and in groups, has been implemented for method assessment, as illustrated in Table \ref{table:compa15}.

To obtain a more well-rounded efficiency evaluation, the PL metric is further considered. As stated in {\ref{table:compa}} and {\ref{table:compa15}} the proposed model MeSA demonstrated shorter trajectories of $18.56$m and $19.69$m, compared to other approaches. CADRL completed paths of $22.17$m, while LSTM-RL, SARL and CAM-RL recorded $21.51$, $19.74$ and $21.78$ meters, respectively. As shown in {\ref{table:compa15}}, similar to the SR and CR, scenarios with more obstacles resulted in a decrease in efficiency for all models. In scenarios with $10$ and $15$ humans, the proposed method MeSA consistently outperformed all other methods by completing tasks in less time and with shorter trajectories.
Conclusively, the integration of memory-enabled elements into a global planning system facilitates action predictions based on current contexts, adapting to complex crowd navigation tasks, and securing a flexible policy that allows the robot to navigate without encountering freezing issues. The simulated experimental results further unveil that the proposed method excels over the other four learning baselines across all metrics by capturing dynamically altering features since it utilizes the acquired awareness as long as necessary.

\begin{table}[!t]
\caption{Quantitative results in a grouped-human scenario with ten humans.}
\label{table:compa}
\begin{center}
\begin{tabular}{l c c c c c c }
\hline
Method & SR & CR & NT & TO & PL & AR \\
\hline
CADRL \cite{chen2017decentralized} & 0.80 & 0.09 & 24.31 & 0.23 & 22.17 & 0.8004 \\

LSTM-RL \cite{everett2018motion} & 0.96 & 0.03 & 21.18 & 0.23 & 21.51 & 1.0404 \\

SARL \cite{chen2019crowd} & 0.97 & 0.03 & 19.73 & 0.22 & 19.74 & 1.3325 \\ 

CAM-RL \cite{samsani2023memory} & 0.93 & 0.04 & 22.51 & 0.19 & 21.78 & 1.0034 \\ 

MeSA-DRL (Proposed) & 0.99 & 0.01 & 18.15 & 0.16 & 18.56 & 1.4529 \\

\hline
\end{tabular}
\end{center}
\end{table} 

\begin{table}[!t]
\caption{Quantitative results in a grouped-human scenario with fifteen humans.}
\label{table:compa15}
\begin{center}
\begin{tabular}{l c c c c c c }
\hline
Method & SR & CR & NT & TO & PL & AR \\
\hline
CADRL \cite{chen2017decentralized} & 0.90 & 0.05 & 20.06 & 0.25 & 23.01 & 1.0021 \\

LSTM-RL \cite{everett2018motion} & 0.95 & 0.04 & 22.15 & 0.17 & 22.95 & 1.1201 \\

SARL \cite{chen2019crowd} & 0.96 & 0.03 & 21.28 & 0.26 & 21.35 & 1.2025  \\ 

CAM-RL \cite{samsani2023memory} & 0.81 & 0.07 & 21.71 & 0.23 & 24.05 & 0.9844 \\ 

MeSA-DRL(Proposed) & 0.99 & 0.01 & 18.87 & 0.17 & 19.69 & 1.4119 \\

\hline
\end{tabular}
\end{center}
\end{table} 


A post hoc analysis employing a Mann-Whitney U-test for pairs is conducted to discern statistical significance between the proposed approach and the comparative baseline methods. All metrics are reported and interpreted at $p_{value} \leq 0.05$ as displayed in Table \ref{statistical_analysis}. Given that the p-value is below the significance level in success rate, collision rate, and average navigation time, it can be deduced that there is evidence to reject the null hypothesis (presuming no disparity across groups) and conclude that the proposed approach is statistically more significant than others.


\begin{table}[!t]
\begin{center}
\caption{Statistical analysis using Mann-Whitney U test.}
\label{statistical_analysis}       
\begin{tabular}{l c c c c}
\hline
 Metrics & U-value & p-value & RBC & CLES \\
\hline
Success Rate & 2.0 & 0.04419 & 0.08 & 0.92\\
Collision Rate & 7.0 & 2.30506e-34 & 0.28 & 0.72\\
Average Navigation Time & 4.0 & 1.97317e-09 & 0.16 & 0.84\\
Timeout & 8.0 & 7.17994e-46 & 0.32 & 0.68\\
Total Reward & 2.0 & 0.04550 & 0.08 & 0.92\\
\hline
\end{tabular}
\end{center}
\end{table}

\textbf{Qualitative Results:} 
The efficacy of the proposed method is qualitatively analyzed through trajectory evaluation, determining the average deviation from a designated ideal path. The Euclidean distance, calculated between each sampled robot position and the ideal path segment, quantifies this deviation, representing the discrepancy between the robot's current position from the optimal path. This methodology is illustrated in Fig. \ref{fig:paths} and Fig. \ref{fig:deviation}, providing a neutral assessment of deviations over diverse path lengths. CADRL noticeably diverges from the ideal path, displaying a higher average deviation, while the LSTM-RL method, although often identifying a route nearer to the desired one, prompts the robot to undertake detours, creating rather erratic paths. In comparison, the trajectory utilizing the SARL method becomes notably irregular and deviating, failing to sustain a straightforward linear trajectory. Although the CAM-RL method leads the robot toward the goal with minimal disturbance, it opts for a route that, due to its substantial deviation from the desired path, is less than ideal. The proposed method ensures smooth navigation for the robot by utilizing a multi-term reward, motivating it to adhere to the optimal path without disruptions, as demonstrated in the trajectory evaluation and its deviation in Fig. \ref{fig:deviation}.
In a fair environment, with fifteen identical human courses, navigation trajectories employing various techniques are contrasted as shown in Fig. \ref{fig:compa15h}.
During navigation, upon encountering pedestrians in dynamic groups, the robot under the guidance of CADRL surpasses them creating unsafe conditions for pedestrians.
With an increase in density, CADRL fails to reach the goal ending with a collision as is illustrated in Fig. \ref{fig:cadrl_15h}.
As seen in Fig. \ref{fig:lstm_15h} the path derived by the LSTM-RL method distinctly showcases that indecisive and aggressive behaviors elevate risk by directing the robot into a cluster of humans.
\begin{figure*}[!t]
\centering
\begin{subfigure}{.23\textwidth}
  \centering
  \includegraphics[width=\textwidth]{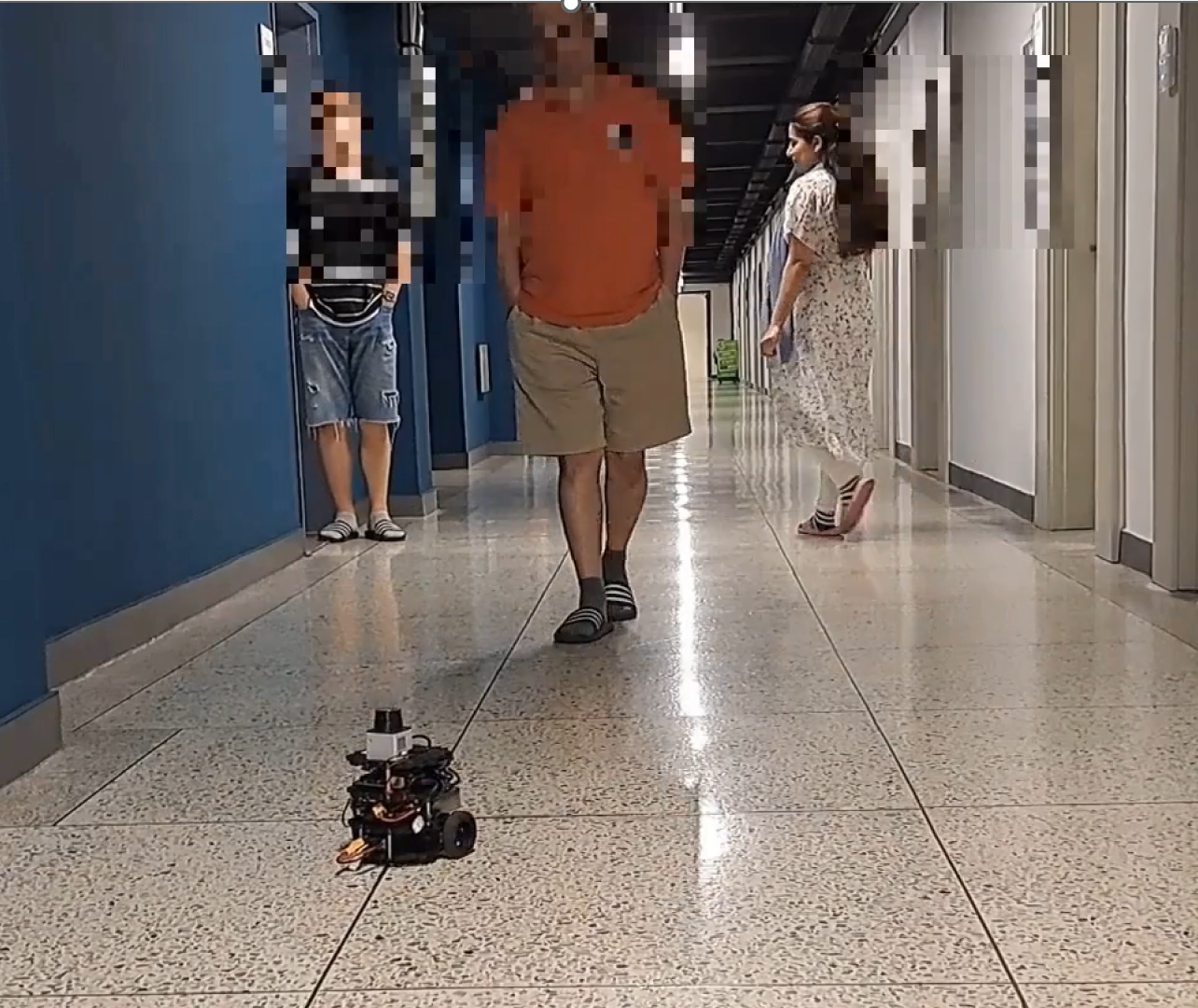}
  \caption{}
\end{subfigure}%
\hspace{0.2cm}
\begin{subfigure}{.25\textwidth}
  \centering
  \includegraphics[width=\textwidth]{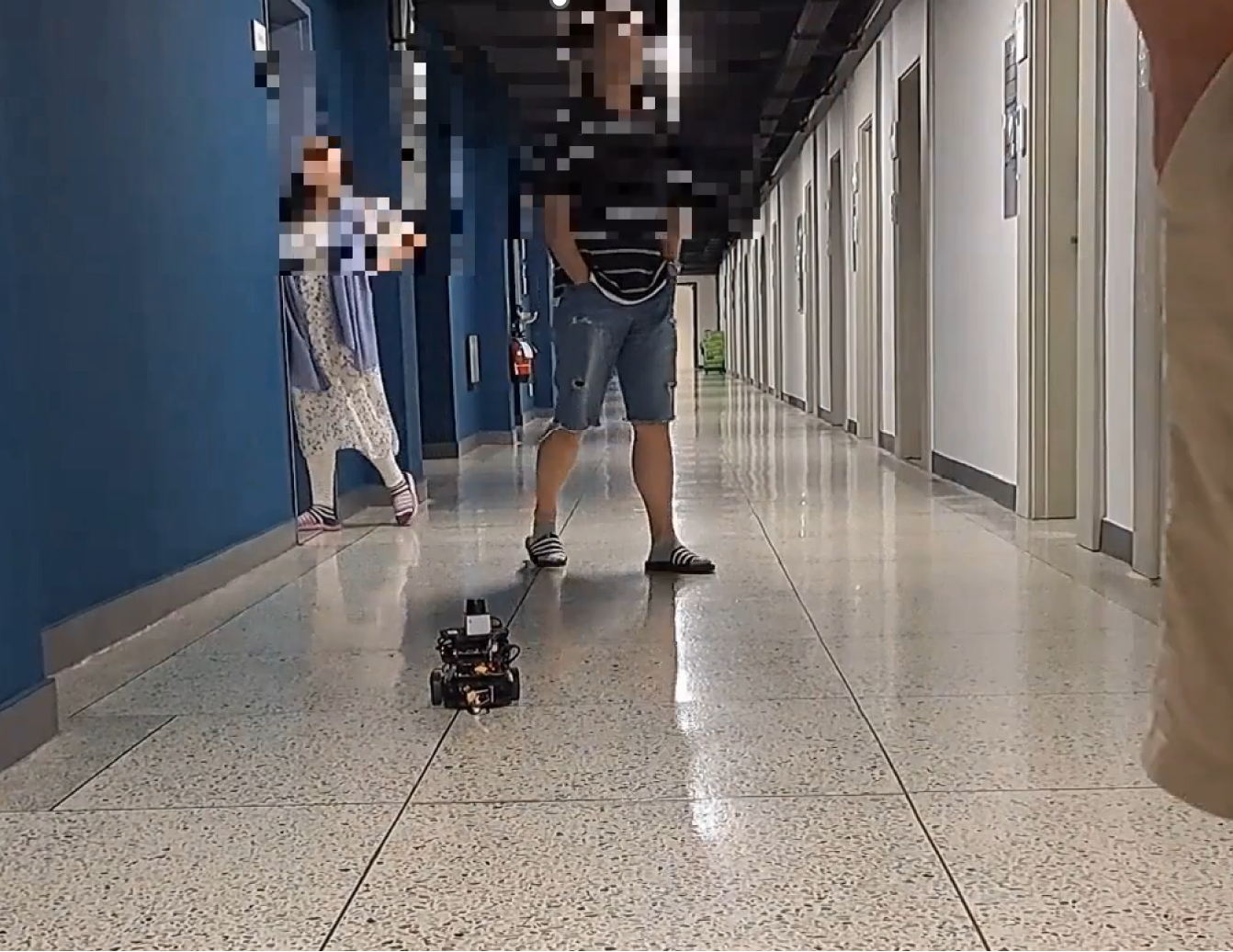}
  \caption{}
\end{subfigure}%
\hspace{0.2cm}
\begin{subfigure}{.23\textwidth}
  \centering
  \includegraphics[width=\textwidth]{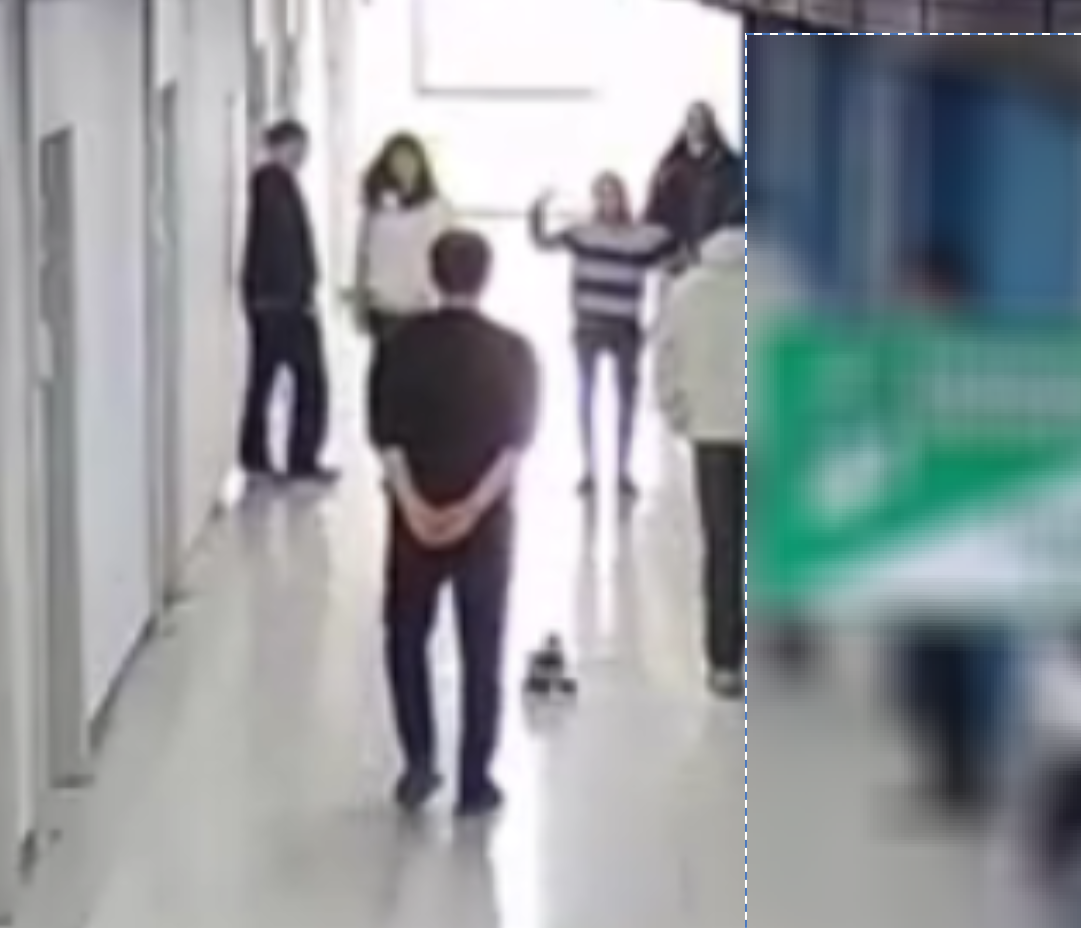}
  \caption{}
\end{subfigure}%
\hfill
\begin{subfigure}{.225\textwidth}
  \centering
  \includegraphics[width=\textwidth]{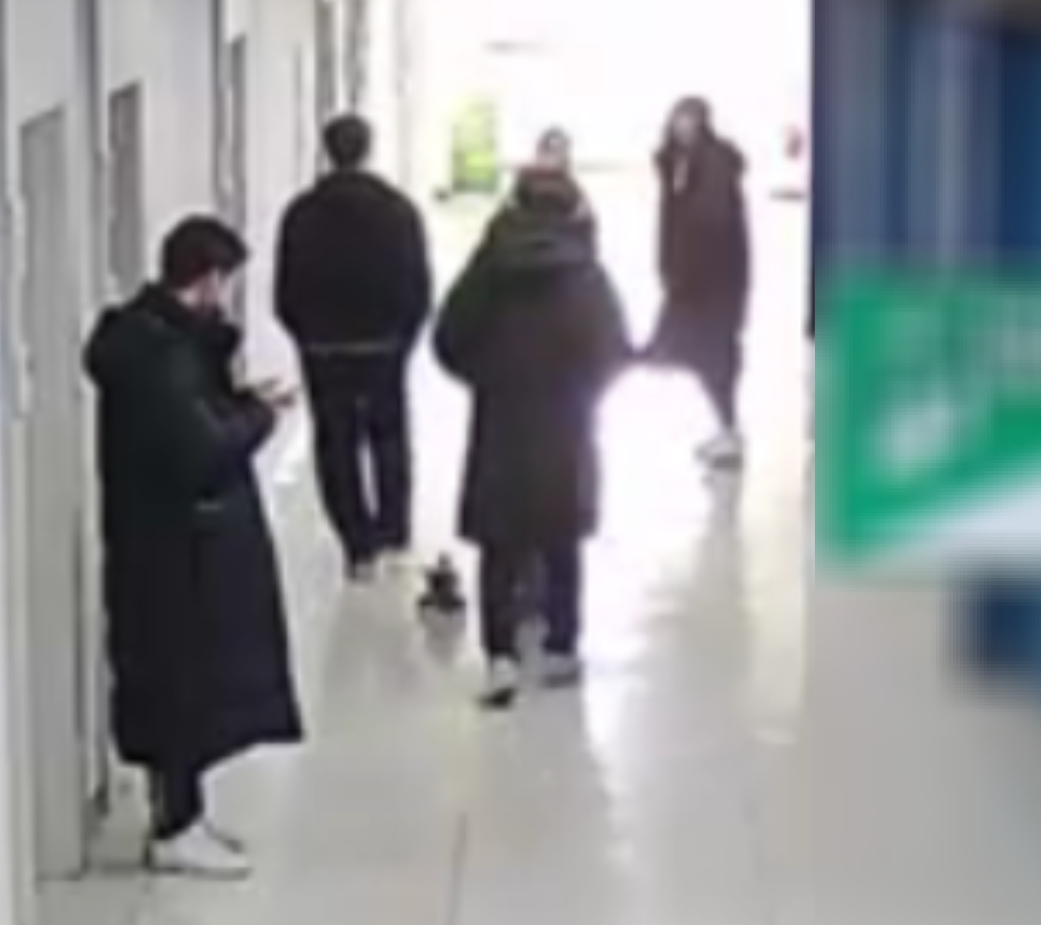}
  \caption{}
\end{subfigure}%
\caption{Real world experiments in crowded scenarios.}
\label{fig:rwexp}
\end{figure*}
The SARL method induces an overly cautious approach, yielding elongated travel times and uneven paths as shown in Fig. \ref{fig:sarl_15h}.
Employing the CAM-RL method, the robot experiences a period of immobility, failing to discern a path leading to the goal.
In scenarios involving 15 humans, the CAM-RL method to avoid a collision changes its direction, freezes again, and waits for the humans to pass, resulting in a longer time to reach the goal, as demonstrated in Fig. \ref{fig:camrl_15h}.
Fig. \ref{fig:our_15h} shows the trajectory of the proposed MeSA-DRL method. Using a memory-enabled framework employing a dynamic warning zone reward, the proposed method excels over baseline methods in ensuring safety in challenging scenarios, thereby diminishing the likelihood of freezing robot scenarios and minimizing arrival time by maintaining a smooth trajectory.

\section{Real-world Experiments}

To evaluate the efficacy of the proposed approach in authentic crowd scenarios, we utilized a Turtlebot 3. Originally equipped with a Raspberry Pi 3, the Turtlebot is connected to a host computer running the Robot Operating System (ROS) on Ubuntu 18.04. The host computer processes real-time inputs regarding the state of humans and the robot, as well as the robot’s motion commands, which are derived from the proposed pre-trained MeSA policy. The Turtlebot 3 can reach a maximum speed of $0.23$m/s. To accommodate the maximum speed of robot, humans are instructed to walk at a slower pace. The robot is equipped with a Hokuyo 2D scanning sensor, which has a detection distance of $5.6$m and offers 360-degree resolution for mapping and obstacle identification. In the proposed method, it is essential for the robot to have prior knowledge of its position and destination in 2D coordinates. Therefore, a 2D occupancy grid map of the environment is created using the GMapping package {\cite{gmapping}}. The precise location of the robot is determined using an odometer, in conjunction with LIDAR and the AMCL package {\cite{amcl}}. Human detection is carried out with the obstacle-detector package {\cite{obstacle_detector}} to estimate the positions and velocities of humans. The robot is deployed in a corridor with various human flow/movement patterns to comprehensively assess the model's adaptability. Despite the inherent challenges, the robot successfully covers a distance of $8$m in merely $40.05$sec. Utilizing the model, the robot demonstrates swift responsiveness to human movements, promptly halting and smoothly redirecting its course as needed. This indicates its adaptability to demanding environments without requiring retraining. Fig. {\ref{fig:rwexp}} shows the real-world experiments in crowded scenarios, demonstrating the effectiveness of the proposed scheme.

\section{CONCLUSIONS}
In this study, a memory-enabled DRL framework is developed for robot navigation to effectively retain information concerning humans in the environment, employing gated recurrent units to model the sequential dependencies. 
Using the acquired knowledge reciprocal human-robot interactions are extracted. 
This method is further integrated into a navigation framework, inclusive of a map and global planner. 
Consequently, the robot achieves the capacity to react rapidly and maintain its trajectory, without experiencing the freezing-robot problem.
Training of the robot is facilitated through a multi-term reward, encouraging safe, collision-free movement towards its intended destination. Results from the research, grounded in various evaluation contexts, indicate that the learned policy enables the robot to attain a higher percentage of successful trials, fewer collisions, and reduced arrival times in comparison to state-of-the-art algorithms. Real-world implementation illustrates the method’s potential to proficiently navigate through complex reciprocal human relationships.

\bibliographystyle{IEEEtran}
\bibliography{main}

\begin{thebibliography}{10}
\providecommand{\url}[1]{#1}
\csname url@samestyle\endcsname
\providecommand{\newblock}{\relax}
\providecommand{\bibinfo}[2]{#2}
\providecommand{\BIBentrySTDinterwordspacing}{\spaceskip=0pt\relax}
\providecommand{\BIBentryALTinterwordstretchfactor}{4}
\providecommand{\BIBentryALTinterwordspacing}{\spaceskip=\fontdimen2\font plus
\BIBentryALTinterwordstretchfactor\fontdimen3\font minus \fontdimen4\font\relax}
\providecommand{\BIBforeignlanguage}[2]{{%
\expandafter\ifx\csname l@#1\endcsname\relax
\typeout{** WARNING: IEEEtran.bst: No hyphenation pattern has been}%
\typeout{** loaded for the language `#1'. Using the pattern for}%
\typeout{** the default language instead.}%
\else
\language=\csname l@#1\endcsname
\fi
#2}}
\providecommand{\BIBdecl}{\relax}
\BIBdecl

\bibitem{kim2021preference}
S.~S. Kim, J.~Kim, F.~Badu-Baiden, M.~Giroux, and Y.~Choi, ``Preference for robot service or human service in hotels? impacts of the covid-19 pandemic,'' \emph{International Journal of Hospitality Management}, vol.~93, p. 102795, 2021.

\bibitem{xiao2022motion}
X.~Xiao, B.~Liu, G.~Warnell, and P.~Stone, ``Motion planning and control for mobile robot navigation using machine learning: a survey,'' \emph{Autonomous Robots}, vol.~46, no.~5, pp. 569--597, 2022.

\bibitem{fan2019getting}
T.~Fan, X.~Cheng, J.~Pan, P.~Long, W.~Liu, R.~Yang, and D.~Manocha, ``Getting robots unfrozen and unlost in dense pedestrian crowds,'' \emph{IEEE Robotics and Automation Letters}, vol.~4, no.~2, pp. 1178--1185, 2019.

\bibitem{bonny2022highly}
T.~Bonny and M.~Kashkash, ``Highly optimized q-learning-based bees approach for mobile robot path planning in static and dynamic environments,'' \emph{Journal of Field Robotics}, vol.~39, no.~4, pp. 317--334, 2022.

\bibitem{han2022reinforcement}
R.~Han, S.~Chen, S.~Wang, Z.~Zhang, R.~Gao, Q.~Hao, and J.~Pan, ``Reinforcement learned distributed multi-robot navigation with reciprocal velocity obstacle shaped rewards,'' \emph{IEEE Robotics and Automation Letters}, vol.~7, no.~3, pp. 5896--5903, 2022.

\bibitem{everett2021collision}
M.~Everett, Y.~F. Chen, and J.~P. How, ``Collision avoidance in pedestrian-rich environments with deep reinforcement learning,'' \emph{IEEE Access}, vol.~9, pp. 10\,357--10\,377, 2021.

\bibitem{zhu2022hierarchical}
W.~Zhu and M.~Hayashibe, ``A hierarchical deep reinforcement learning framework with high efficiency and generalization for fast and safe navigation,'' \emph{IEEE Transactions on Industrial Electronics}, vol.~70, no.~5, pp. 4962--4971, 2022.

\bibitem{cimurs2021goal}
R.~Cimurs, I.~H. Suh, and J.~H. Lee, ``Goal-driven autonomous exploration through deep reinforcement learning,'' \emph{IEEE Robotics and Automation Letters}, vol.~7, no.~2, pp. 730--737, 2021.

\bibitem{chen2017decentralized}
Y.~F. Chen, M.~Liu, M.~Everett, and J.~P. How, ``Decentralized non-communicating multiagent collision avoidance with deep reinforcement learning,'' in \emph{2017 IEEE international conference on robotics and automation (ICRA)}.\hskip 1em plus 0.5em minus 0.4em\relax IEEE, 2017, pp. 285--292.

\bibitem{everett2018motion}
M.~Everett, Y.~F. Chen, and J.~P. How, ``Motion planning among dynamic, decision-making agents with deep reinforcement learning,'' in \emph{2018 IEEE/RSJ International Conference on Intelligent Robots and Systems (IROS)}.\hskip 1em plus 0.5em minus 0.4em\relax IEEE, 2018, pp. 3052--3059.

\bibitem{chen2019crowd}
C.~Chen, Y.~Liu, S.~Kreiss, and A.~Alahi, ``Crowd-robot interaction: Crowd-aware robot navigation with attention-based deep reinforcement learning,'' in \emph{2019 International Conference on Robotics and Automation (ICRA)}.\hskip 1em plus 0.5em minus 0.4em\relax IEEE, 2019, pp. 6015--6022.

\bibitem{liu2021decentralized}
S.~Liu, P.~Chang, W.~Liang, N.~Chakraborty, and K.~Driggs-Campbell, ``Decentralized structural-rnn for robot crowd navigation with deep reinforcement learning,'' in \emph{2021 IEEE International Conference on Robotics and Automation (ICRA)}.\hskip 1em plus 0.5em minus 0.4em\relax IEEE, 2021, pp. 3517--3524.

\bibitem{singh2023behavior}
K.~J. Singh, D.~S. Kapoor, M.~Abouhawwash, J.~F. Al-Amri, S.~Mahajan, and A.~K. Pandit, ``Behavior of delivery robot in human-robot collaborative spaces during navigation.'' \emph{Intelligent Automation \& Soft Computing}, vol.~35, no.~1, 2023.

\bibitem{reich2020memory}
G.~M. Reich, M.~Antoniou, and C.~J. Baker, ``Memory-enhanced cognitive radar for autonomous navigation,'' \emph{IET Radar, Sonar \& Navigation}, vol.~14, no.~9, pp. 1287--1296, 2020.

\bibitem{van2008reciprocal}
J.~Van~den Berg, M.~Lin, and D.~Manocha, ``Reciprocal velocity obstacles for real-time multi-agent navigation,'' in \emph{2008 IEEE international conference on robotics and automation}.\hskip 1em plus 0.5em minus 0.4em\relax Ieee, 2008, pp. 1928--1935.

\bibitem{van2011reciprocal}
J.~Van Den~Berg, S.~J. Guy, M.~Lin, and D.~Manocha, ``Reciprocal n-body collision avoidance,'' in \emph{Robotics Research: The 14th International Symposium ISRR}.\hskip 1em plus 0.5em minus 0.4em\relax Springer, 2011, pp. 3--19.

\bibitem{ferrer2013robot}
G.~Ferrer, A.~Garrell, and A.~Sanfeliu, ``Robot companion: A social-force based approach with human awareness-navigation in crowded environments,'' in \emph{2013 IEEE/RSJ International Conference on Intelligent Robots and Systems}.\hskip 1em plus 0.5em minus 0.4em\relax IEEE, 2013, pp. 1688--1694.

\bibitem{fan2020distributed}
T.~Fan, P.~Long, W.~Liu, and J.~Pan, ``Distributed multi-robot collision avoidance via deep reinforcement learning for navigation in complex scenarios,'' \emph{The International Journal of Robotics Research}, vol.~39, no.~7, pp. 856--892, 2020.

\bibitem{perez2021robot}
C.~P{\'e}rez-D’Arpino, C.~Liu, P.~Goebel, R.~Mart{\'\i}n-Mart{\'\i}n, and S.~Savarese, ``Robot navigation in constrained pedestrian environments using reinforcement learning,'' in \emph{2021 IEEE International Conference on Robotics and Automation (ICRA)}.\hskip 1em plus 0.5em minus 0.4em\relax IEEE, 2021, pp. 1140--1146.

\bibitem{sathyamoorthy2020frozone}
A.~J. Sathyamoorthy, U.~Patel, T.~Guan, and D.~Manocha, ``Frozone: Freezing-free, pedestrian-friendly navigation in human crowds,'' \emph{IEEE Robotics and Automation Letters}, vol.~5, no.~3, pp. 4352--4359, 2020.

\bibitem{jin2020mapless}
J.~Jin, N.~M. Nguyen, N.~Sakib, D.~Graves, H.~Yao, and M.~Jagersand, ``Mapless navigation among dynamics with social-safety-awareness: a reinforcement learning approach from 2d laser scans,'' in \emph{2020 IEEE international conference on robotics and automation (ICRA)}.\hskip 1em plus 0.5em minus 0.4em\relax IEEE, 2020, pp. 6979--6985.

\bibitem{chen2020robot}
Y.~Chen, C.~Liu, B.~E. Shi, and M.~Liu, ``Robot navigation in crowds by graph convolutional networks with attention learned from human gaze,'' \emph{IEEE Robotics and Automation Letters}, vol.~5, no.~2, pp. 2754--2761, 2020.

\bibitem{zhu2022collision}
K.~Zhu, B.~Li, W.~Zhe, and T.~Zhang, ``Collision avoidance among dense heterogeneous agents using deep reinforcement learning,'' \emph{IEEE Robotics and Automation Letters}, vol.~8, no.~1, pp. 57--64, 2022.

\bibitem{samsani2021socially}
S.~S. Samsani and M.~S. Muhammad, ``Socially compliant robot navigation in crowded environment by human behavior resemblance using deep reinforcement learning,'' \emph{IEEE Robotics and Automation Letters}, vol.~6, no.~3, pp. 5223--5230, 2021.

\bibitem{montero2023dynamic}
E.~E. Montero, H.~Mutahira, N.~Pico, and M.~S. Muhammad, ``Dynamic warning zone and a short-distance goal for autonomous robot navigation using deep reinforcement learning,'' \emph{Complex \& Intelligent Systems}, pp. 1--18, 2023.

\bibitem{yuan2019novel}
J.~Yuan, H.~Wang, C.~Lin, D.~Liu, and D.~Yu, ``A novel gru-rnn network model for dynamic path planning of mobile robot,'' \emph{IEEE Access}, vol.~7, pp. 15\,140--15\,151, 2019.

\bibitem{zeng2019navigation}
J.~Zeng, R.~Ju, L.~Qin, Y.~Hu, Q.~Yin, and C.~Hu, ``Navigation in unknown dynamic environments based on deep reinforcement learning,'' \emph{Sensors}, vol.~19, no.~18, p. 3837, 2019.

\bibitem{cho2014properties}
K.~Cho, B.~Van~Merri{\"e}nboer, D.~Bahdanau, and Y.~Bengio, ``On the properties of neural machine translation: Encoder-decoder approaches,'' \emph{arXiv preprint arXiv:1409.1259}, 2014.

\bibitem{chung2014empirical}
J.~Chung, C.~Gulcehre, K.~Cho, and Y.~Bengio, ``Empirical evaluation of gated recurrent neural networks on sequence modeling,'' \emph{arXiv preprint arXiv:1412.3555}, 2014.

\bibitem{paszke2019pytorch}
A.~Paszke, S.~Gross, F.~Massa, A.~Lerer, J.~Bradbury, G.~Chanan, T.~Killeen, Z.~Lin, N.~Gimelshein, L.~Antiga \emph{et~al.}, ``Pytorch: An imperative style, high-performance deep learning library,'' \emph{Advances in neural information processing systems}, vol.~32, 2019.

\bibitem{samsani2023memory}
S.~S. Samsani, H.~Mutahira, and M.~S. Muhammad, ``Memory-based crowd-aware robot navigation using deep reinforcement learning,'' \emph{Complex \& Intelligent Systems}, vol.~9, no.~2, pp. 2147--2158, 2023.

\bibitem{gmapping}
B.~Gerkey, ``{Gmapping ros package.}'' \url{https://wiki.ros.org/gmapping/}, 2021, accessed: 2023-10-17.

\bibitem{amcl}
Gerkey, ``{Amcl ros package.}'' \url{https://wiki.ros.org/amcl/}, 2022, accessed: 2023-10-17.

\bibitem{obstacle_detector}
M.~Przybyla, ``Obstacle detector ros package.'' \url{https://github.com/tysik/obstacle detector}, 2022, accessed: 2023-10-17.

\end{thebibliography}

\end{document}